\documentclass[letterpaper, 10pt, conference]{ieeeconf}
\usepackage{amsmath} 
\usepackage{amssymb}  
\usepackage{graphicx}
\usepackage{caption}
\usepackage{subcaption}
\usepackage{multirow}
\usepackage{amsfonts}
\usepackage{tabularx}
\usepackage{amsfonts}
\usepackage{algorithm,algpseudocode}
\usepackage{bm}
\usepackage{enumerate}
\usepackage{verbatim}
\usepackage{float}
\usepackage{siunitx}
\usepackage{mdframed}
\usepackage{adjustbox}
\usepackage{authblk}
\usepackage{color}
\usepackage{verbatim}
\usepackage{float}
\usepackage{siunitx}
\usepackage{mdframed}
\usepackage{booktabs}
\makeatletter
\let\NAT@parse\undefined
\makeatother
\usepackage{hyperref}
\usepackage{cleveref}
\newcommand{\specialcell}[2][c]{%
  \begin{tabular}[#1]{@{}c@{}}#2\end{tabular}}

\title{\LARGE \bf PENet: A Joint Panoptic Edge Detection Network}

\author{Yang Zhou and Giuseppe Loianno
\thanks{$^*$These authors contributed equally.}
\thanks{The authors are with the New York University, Tandon School of Engineering, Brooklyn, NY 11201, USA. {\tt\footnotesize email: \{yangzhou, loiannog\}@nyu.edu}.}
}

\begin{document}
\maketitle

\thispagestyle{empty}
\pagestyle{empty}

\begin{abstract}

In recent years, compact and efficient scene understanding representations have gained popularity in increasing situational awareness and autonomy of robotic systems. In this work, we illustrate the concept of a panoptic edge segmentation and propose PENet, a novel detection network called that combines semantic edge detection and instance-level perception into a compact panoptic edge representation. This is obtained through a joint network by multi-task learning that concurrently predicts semantic edges, instance centers and
offset flow map without bounding box predictions exploiting the cross-task correlations among the tasks. The proposed approach allows extending semantic edge detection to panoptic edge detection which encapsulates both category-aware and instance-aware segmentation. We validate the proposed panoptic edge segmentation method and demonstrate its effectiveness on the real-world Cityscapes dataset.

\end{abstract}

\section*{Supplementary Material}
 \textbf{Codes}: \href{https://t.ly/U3sDD}{https://t.ly/U3sDD}
 
 \textbf{Video}: \href{https://t.ly/DlzX}{https://t.ly/DlzX}

\vspace{-5pt}
\section{Introduction}~\label{sec:introduction}
Semantic scene understanding~\cite{garg2020semantics} is an important task in different areas such as autonomous driving, search and rescue, and robot navigation. A compact environmental representation with geometric and semantic information can raise robots' situational awareness and assist them execute complex autonomous tasks. Camera sensors are widely used in the robotics field for semantic scene understanding due to their low cost, ability to obtain semantic information from the appearance and textures of the scene, and compatibility with Size, Weight, and Power (SWaP) constraints of small-scale robots. 

The recent trend of deep learning has produced different semantic representations in computer vision and robotics. Researchers' interests span from object detection~\cite{ren2015faster}, semantic segmentation~\cite{long2015fully}, panoptic segmentation~\cite{kirillov2019panoptic} and scene understanding~\cite{xiao2018unified}. Semantic edge which was first proposed in CASENet~\cite{Yu2017CASENet:Detection} classifies each edge pixel with corresponding semantic categories. It offers a unique representation that jointly combines geometric and semantic information. Other works including~\cite{yu2018simultaneous}, \cite{acuna2019devil} and \cite{hu2019dynamic} improve the accuracy of semantic edge detection by proposing different alignment mechanisms and feature fusion strategies. Our previous work FENet~\cite{zhou2020fenet} proposes the first real-time semantic edge network for embedded robotic systems with SWaP constraints. The semantic edge provides a compact semantic representation that is memory-efficient for communication and suitable for real-time semantic feature extraction.
\begin{figure}
    \centering
    \includegraphics[width=0.9\linewidth]{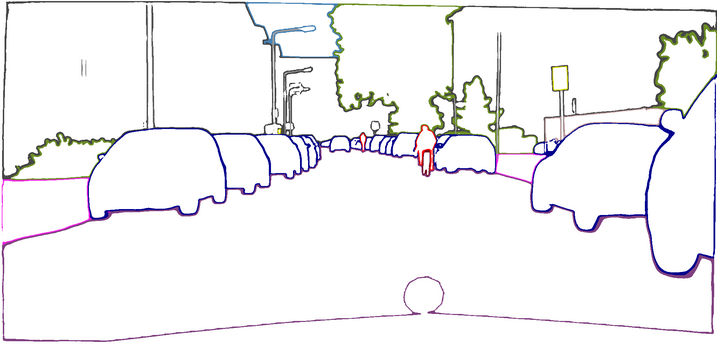}
    \includegraphics[width=0.9\linewidth]{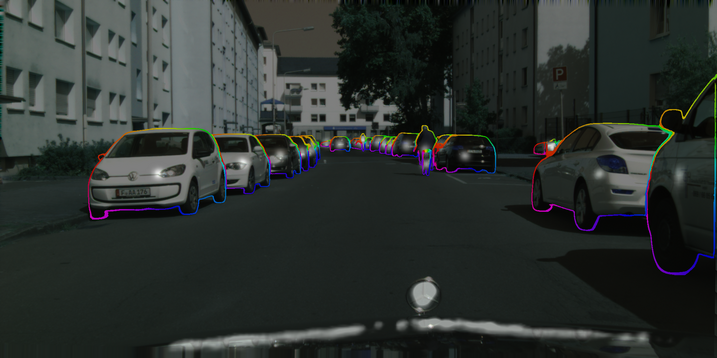}
    \includegraphics[width=0.9\linewidth]{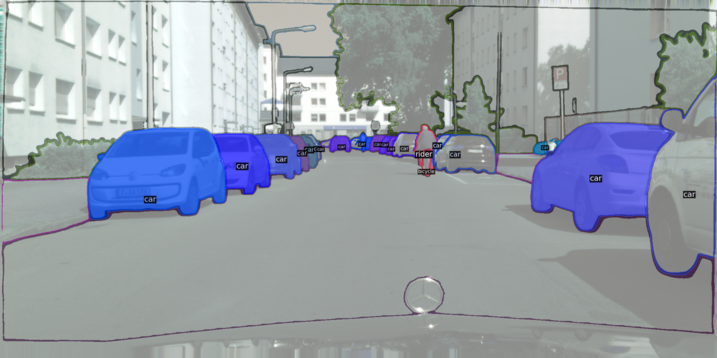}
    \caption{Combining the semantic edge detection (top) and Instance center and offset regression (center), our proposed PENet (bottom) can obtain panoptic edges that differentiate different categories and instances of edges. We fill the closed instance edge contours for visualization purpose.}
    \label{fig:teaser}
    \vspace{-15pt}
\end{figure}

However, existing semantic edge representation cannot support instance-level semantic perception, which is often required by downstream robotics tasks, including planning, mapping, and navigation. In addition to category-level edge pixel classification, it is also important to categorize edges belonging to different instances. For example, in autonomous driving scenarios, we would like to track different instances of vehicles and distinguish them from each other. 
In Fig.~\ref{fig:teaser}, compared to semantic edge detection, panoptic edge detection can differentiate different vehicle instances that belong to the same category.

To address this issue, in this work, we propose a novel approach that combines instance information with semantic edge representation. In~\cite{hu2019panoptic}, the authors propose the first panoptic edge detection method. They combine the information from object detection and semantic edge detection branches respectively. Both branches share the same feature encoder. By fusing the information of the output from both tasks, the final panoptic edge detection result is obtained. 

Compared to~\cite{hu2019dynamic}, our approach reformulates the semantic edge definition from multi-label per pixel to single-label per pixel, in order to address the issue of inter-class instance edge. We formulate the panoptic upon the new semantic edge definition. Our method does not rely on bounding-box based object detection. Conversely, we employ one sub-network to predict the semantic edges, and an additional sub-network to predict the edge category, the center, and edge pixels' offset from their corresponding centers. Finally, we employ a clustering technique to fuse the center and offset predictions with semantic edge detection to predict the final panoptic edge detection result. Our proposed approach shares a joint encoder for all sub-networks producing a feature embedding that is memory friendly and efficient. 

This paper presents several contributions. First, we provide a formulation of panoptic edge since this is not straightforward as for other types of segmentation tasks.
Second, we propose a novel panoptic edge detection multi-task learning design that jointly predicts semantic edges, instance centers, and offset without bounding box predictions exploiting the cross-task correlations. Finally, we demonstrate and validate the performance of our approach in real-world settings using the Cityscapes~\cite{cordts2016cityscapes} dataset, and we open source our implementation.
\vspace{-5pt}
\section{Related Works}~\label{sec:relatedworks}
\textbf{Panoptic Segmentation}.
This segmentation task can generate a scene representation that contains the segmentation mask of "stuff" (non-instance such as sky and road) and "thing" (such as cars and humans which we need to identify different instances), therefore, unifying  semantic segmentation and instance segmentation. Panoptic Segmentation also separates objects' instances with different instance ID~\cite{kirillov2019panoptic}. UPSNet~\cite{xiong2019upsnet} proposes a unified panoptic segmentation network which combines a semantic segmentation head with a Mask R-CNN~\cite{he2017mask} style instance segmentation head and introduces a parameter-free panoptic head to predict the panoptic label. Panoptic FPN~\cite{kirillov2019panoptica} also uses a semantic segmentation branch and a Mask R-CNN instance segmentation branch, but it uses a shared Feature Pyramid Network network as the backbone.
FPSNet~\cite{de2020fast} solves the same problem as a dense pixel-wise classification problem,  which predicts a class label or an instance ID for each pixel, which runs faster than the above methods.
Panoptic-DeepLab~\cite{cheng2020panoptic} uses a class-agnostic instance segmentation branch that includes a simple instance center regression, which improves the accuracy of panoptic segmentation with a simple and fast design. 

Our work is inspired by the design choice of Panoptic-DeepLab. However, rather than separating the prediction of the instance edges and semantic edges, we obtain the instance edges by combining the prediction of semantic edges, the center prediction of instances, and center regression of the instances. Finally, we fuse the semantic edges with the instance edges to produce the panoptic edges.

\textbf{Semantic Edge Detection}.
This type of segmentation is different compared to semantic segmentation. Semantic segmentation predicts the category of the pixel inside the target, while semantic edge only focuses on the boundary of the target and predicts the multi-label categories of the boundary pixels. 
CASENet~\cite{Yu2017CASENet:Detection} is the first approach to propose the semantic edge detection method. It predicts edge pixels and classifies the category of each edge pixel. Notably, each edge pixel can belong to multiple categories. Later, SEAL~\cite{yu2018simultaneous} and STEAL~\cite{acuna2019devil} improve the quality of the semantic edge detection by aligning edges using latent variable optimization and reasoning the annotation noise. DFF~\cite{hu2019dynamic} adaptively adjusts the weights of feature maps from different levels according to the high-level feature map, which further boosts up the accuracy of semantic edge detection. More recently, in order to deploy real-time semantic edge detection on edge computing devices, our FENet work~\cite{zhou2020fenet} proposes the first real-time semantic edge detection method which can run on an embedded robotic platform at more than $30$ Hz.  

This work extends semantic edge detection to panoptic edge detection by introducing center regression and head regression. We adopt a fusion strategy to fuse the object center, object offset, and semantic labels to produce the panoptic edges.  In this area, the closest related work PEN~\cite{hu2019panoptic} cannot directly regress the instance edge pixel and employs the bounding-box of the object detection to solve this task. The approach simply combines object detection with semantic edge detection to coarsely produce panoptic edges without properly exploiting cross-task correlations.

\begin{figure*}[ht!]
    \centering
    \includegraphics[width=0.9\linewidth]{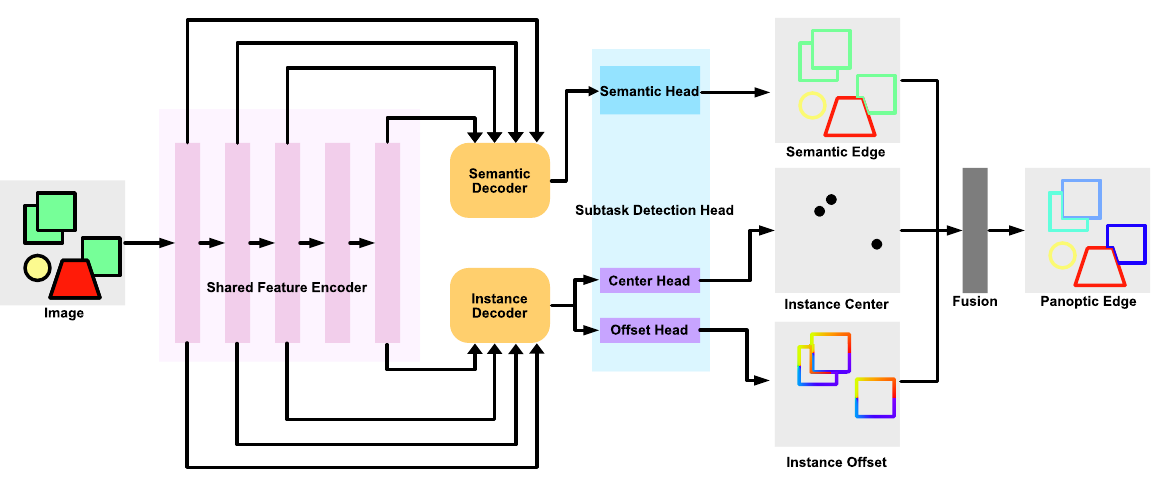}
    \caption{The proposed network architecture. The Shared Feature Encoder generates multi-level features, the Semantic Decoder and Instance Decoder aggregate the input features to decode features for subtask heads, including the semantic head which outputs semantic edges, the center head which outputs instance center regression and the offset head which outputs instance offset regression. Finally, a Fusion module fuses them together to obtain the final Panoptic Edge.}
    \label{fig:pipeline}
    \vspace*{-15pt}
\end{figure*}

\section{Methodology}~\label{sec:methodology}
\vspace{-20pt}
\subsection{Panoptic Edge Formulation}
CASENet~\cite{Yu2017CASENet:Detection} formulates category-aware semantic edge such that each semantic edge pixel can map to multiple labels. 
However, if we want to extend the formulation of semantic edge to panoptic edge segmentation, we need to ensure that each edge pixel of two overlapping instances with different semantic category labels should only reflect a single category label, so that the instance labels of these edges do not have ambiguity. This is reflected in our network design and shown in Section~\ref{sec:head}.
Specifically, we first reformulate the semantic edge by adding a constraint in the network design such that each semantic edge pixel only maps to a single category label. We split semantic edge pixels to two sets: "thing" edge pixels and "stuff" edge pixels similar to the definition of Panoptic Segmentation~\cite{kirillov2019panoptic} based on their semantic categories. To be noticed, we use the concept of "thing" edge and instance edge interchangeably. "thing" edge pixels of the same category can have different instances (e.g., contours of cars), whereas "stuff" edge pixels do not have different instances (e.g., contours of the sky). 

Therefore, we map each semantic edge pixel $p_i$ to a single semantic category label ($S_i \in 1,\cdots K$) and a single instance ID (${ID}_i\in \mathbb{N}_0$): 
$f: p_i \to S_i,~g: p_i \to {ID}_i,~S_i \in 1\cdots K$ where: 1) $K$ is the total number of semantic categories; 2) $\forall g(p_i) < D$, where $D\in\mathbb{N}$ is set to be a large enough integer and larger than the largest instance ID among all semantic category label; 3) $\|\{g(p_i)\}\| = \| \{p_i\}\|,~\forall p_i \text{ s.t. }f(p_i) = S$ which means all instance IDs of edge pixels that belong to the same category have unique instance ID among their own set.
We define the panoptic edge label for each "stuff" pixel and "thing" pixel as $S_i*D +{ID}_i$. Each panoptic edge label is unique among all edge segments. Since "stuff" edges do not have different instances, without loss of generality, we can map all "stuff" edge pixels of same category $S$ to any given instance ID  where $S$ is a "stuff" category. 

\subsection{Panoptic Edge Detection Model Overview}
We predict the panoptic edge by predicting the instance ID and semantic label of each pixel. We call these two tasks instance edge prediction and semantic edge prediction. 
We illustrate the proposed model architecture in Fig.~\ref{fig:pipeline}. 

We use a shared feature encoder for two sub-networks of semantic edge prediction and instance edge prediction. Assume the input image has dimension $3\times H \times W$. 
The sub-network of semantic edge prediction outputs a probability map with dimension of $(K+1)\times H \times W$, where $K$ is the number of pre-defined semantic labels. This probability map predicts the probabilities of $K$ labels and the non-edge labels for each pixel. The instance edge sub-network  outputs the instance center regression and instance offset regression. The instance center regression outputs the geometric center of each instance edge with the dimension of $1 \times H \times W$. The instance offset regression outputs the offset vector maps with dimensions of $2 \times H \times W$ from the current edge pixels to their instance center. Given instance centers and corresponding instance offsets, we can compute the instance IDs for edge pixels.

In the following sections, we detail our proposed framework. We introduce the shared feature encoder in Section~\ref{sec:encoder}.
We show the design of semantic decoder and instance decoder in Section~\ref{sec:decoder}. We also show the panoptic edge fusion strategy in Section~\ref{sec:fusion}. Finally, we show the loss function to train our proposed network in Section~\ref{sec:loss}.

\subsection{Shared Feature Encoder}\label{sec:encoder}
The shared feature encoder (see Fig.~\ref{fig:pipeline}) module uses a backbone network to produce multi-level feature maps for all three subtasks: the semantic edge detection, the instance center regression and the instance offset regression. The Shared Feature Encoder outputs feature maps from different levels to the decoders. Low-level feature maps provide geometric details of the edges, whereas high-level feature maps provide semantic clues for category classification.

We use ResNet~\cite{he2016deep} as the backbone network with a replacement of the first block as three $3\times3$ convolution layers which first layer has stride $2$ followed by a $3\times 3$ max-pooling with stride $2$. Thus, the spatial resolution of the feature maps after the first block is $\frac{H}{4} \times \frac{W}{4}$. 

\subsection{Semantic Decoder and Instance Decoder}\label{sec:decoder}
\begin{figure}[t]
    \centering
    \includegraphics[width=0.7\linewidth]{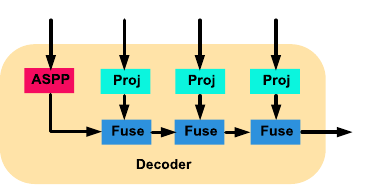}
    \caption{The Decoder Module fuses features from different levels of encoder using ASPP, Projection and Fuse module. }
    \vspace{-5pt}
    \label{fig:decoder}
\end{figure}

The semantic decoder and instance decoder take inputs from the different levels of Shared Feature Encoder, fuse the information across each level, and output the decoded features for their subtasks. 
The output of the semantic decoder will be used by semantic head to produce the semantic edge map. 
The output of the instance decoder will be utilized by center head and offset head to generate instance center map and instance offset map respectively. 
Both decoders share the same meta-architecture illustrated in Fig.~\ref{fig:decoder}.

The features of last encoder block pass into a Astrous Spatial Pyramid Pooling (ASPP) module~\cite{Chen2017RethinkingAC}.
The features of the three additional encoder blocks pass into a Projection module. The Projection module project the features to desired projection channel sizes. We experiment with two different Projection modules. The basic one is a $1\times 1$ convolution layer that projects the input feature channels to smaller values. We call the basic Projection module Proj-A. The advanced one, called Proj-B, is a self-attention module powered by Criss-Cross Attention (CC)~\cite{huang2019ccnet}. The CC module applies axis-wise attention for each element in the feature maps. Then it adds the input feature maps as the residual term to obtain the self-attention output of the feature map. Applying Criss-Cross Attention recursively two times can aggregate the information of all pixels to each pixel.
Compared to non-local attention, which requires an attention map with the same spatial resolution as the input feature map, Criss-Cross Attention is memory efficient and computation-efficient. 
This advanced module first projects the features to projection channel sizes using the convolution layer. Subsequently, it applies the Criss-Cross Attention recursively two times. Finally, it applies another convolution layer.

\begin{figure}[ht!]
\vspace{-10pt}
    \centering
    \includegraphics[width=0.7\linewidth]{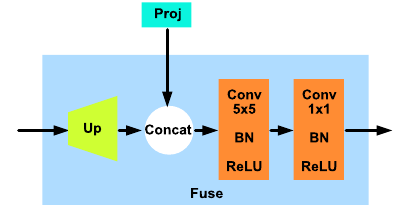}
    \caption{The Fuse module combines low-level and high-level features by upsampling and convolutions.}
    \label{fig:fuse}
    \vspace{-15pt}
\end{figure}

The fuse module shown in Fig.~\ref{fig:fuse} upsamples the low-resolution features and concatenates them with the projected features with high-resolution. The Fuse module then uses a depthwise separable convolution with a $5\times 5$ convolution layer and a $1\times 1$ convolution layer to fuse the features. 

\begin{table*}[htp]
    \centering
    \smallskip
    \caption{Results for different settings of Reweighted loss, ResNet Depth, Pretrain method, Projection Channels and Projection Module Type. The \textbf{bold} value shows the best result and the \underline{underline} value shows the second-best result.}
    \begin{tabular}{ccccccccccccc}
    \toprule
          Model& \specialcell{Reweighted\\Loss} & \specialcell{ResNet\\Depth} & \specialcell{Pretrain} & \specialcell{Projection\\Channnels}  & \specialcell{Projection\\Module}  & PQ $\uparrow$ & SQ $\uparrow$& RQ $\uparrow$  & FLOPS(T)$\downarrow$ & \specialcell{Params\\Size(MiB)} $\downarrow$ \\\midrule
         PENet-A &              & 50 & I &$[16, 32, 64]$ & A & 17.659           & 27.172             & 61.282             & 0.433            & 30.376 \\
         PENet-B & $\checkmark$ & 50 & I &$[16, 32, 64]$ & A & 20.557           & 27.714             & 72.689             & 0.433            & 30.376\\
         PENet-C & $\checkmark$ & 50 & I &$[ 1,  1,  1]$ & A & 20.617           & 27.385             &\textbf{74.131}     & 0.425            & \underline{30.243} \\
         PENet-D & $\checkmark$ & 50 & I &$[16, 32, 64]$ & B &\underline{20.889}& \underline{28.007} &\underline{73.341}  & 0.420            & 30.272\\
         PENet-E & $\checkmark$ & 50 & I + SE &$[16, 32, 64]$ & B&\textbf{21.047}    & \textbf{28.396}    & 72.322             &\underline{0.420} & 30.272\\
         PENet-S & $\checkmark$ & 18 & - &$[ 1,  1,  1]$ & A & 18.887           & 26.447             & 69.277             & \textbf{0.286}   & \textbf{20.700}\\
        \bottomrule
    \end{tabular}
    \label{tab:main}
    \vspace{-14pt}
\end{table*}

\begin{table}[htp]
    \centering
    \smallskip
    \small
    \caption{Detailed PQ evaluation of "thing" and "stuff" edges. The \textbf{bold} value shows the best result and the \underline{underline} value shows the second-best result.}
    \begin{tabular}{c c c c c c c}
    \toprule
    Model & $\text{PQ}_{\text{th}}$ & $\text{SQ}_{\text{th}}$ & $\text{RQ}_{\text{th}}$ & $\text{PQ}_{\text{st}}$ & $\text{SQ}_{\text{st}}$ & $\text{RQ}_{\text{st}}$ \\ \midrule
    A & \textbf{17.33} & 26.40 & \textbf{73.47} & 19.47 & 30.13 & 59.10\\
    B & 16.74 & 26.65 & 70.61 & 24.85 & \underline{30.91} & 80.62 \\
    C & 16.95 & 26.26 & \underline{72.84} & 24.82 & 30.59 & \textbf{81.69} \\
    D & \underline{17.25} & \textbf{26.91} & 72.06 & \underline{25.12} & 31.25 & 80.82 \\
    E & 16.33 & \underline{26.87} & 68.17 & \textbf{25.96} & \textbf{31.95} & \underline{81.54} \\
    S & 14.51 & 25.40 & 64.41 & 23.39 & 29.52 & 78.67 \\
    \bottomrule
    \end{tabular}
    \label{tab:sep_pq}
    \vspace{-10pt}
\end{table}
\subsection{Subtask Detection Head}\label{sec:head}
The subtask detection heads including semantic head, center head and offset head predict the subtask output using aggregated features from the semantic decoders and instance decoders. The format of these subtasks are illustrated in Fig.~\ref{fig:pipeline}. We use square to demonstrate the same-category "thing" edges, and others are "stuff" edges. All subtask detection heads consist of three convolution layers with kernel size $3\times 3$. The semantic head decodes the feature into the $(K+1)\times H \times W$ semantic label map, which predicts the probability of the semantic edge category. The center head decodes the feature into $1\times H \times W$ gaussian heat map, which predicts the possible center locations of "thing" edges. The offset head decodes the feature into a  $2\times H \times W$ feature map, which represents the instance center offset vector from semantic edge pixel to its instance center of "thing" edges.

\subsection{Panoptic Edge Fusion Module}\label{sec:fusion}
The panoptic edge fusion module fuses the semantic edges, the instance centers and the instance offsets to produce the panoptic edge label. 
We first do a clustering step for the predicted instance center since the predicted heatmap of the instance center contains noises.
Given semantic labels of edge pixels, We determine the instance center of an instance edge pixel that has a corresponding semantic edge label according to 1) The distance between the instance center and the pixel; 2) The instance offset vector from the pixel to the instance center.
We generate a vector according to the direction of the normalized offset vector and the magnitude of the distance from the pixel to the predicted instance center. Then, we identify the nearest predicted instance center to the endpoint of the vector. This ensures that each instance edge is associated with a distinct instance ID.
Finally, we merge the semantic edge prediction and instance edge prediction to obtain the panoptic edge prediction.

\subsection{Loss Function}\label{sec:loss}
Our problem is a multi-task learning problem. Therefore, the loss is composed of three parts, one for each subtasks.
For semantic edge detection, inspired by CASENet~\cite{Yu2017CASENet:Detection} and DUpsampling~\cite{tian2019decoders}, we combine $\mathrm{Softmax(\cdot)}$ with temperature $T$ and reweighted cross entropy loss to obtain the semantic edge prediction loss $L_{s}$. We first apply an adaptive-temperature on the multi-label prediction, then we re-weight loss of edge and non-edge pixels using the percentage ratio of them. Compared to conventional $\mathrm{Softmax(\cdot)}$, the adaptive-temperature $\mathrm{Softmax(\cdot)}$ scales down the input by an adaptive trainable temperature $T$
\vspace{-5pt}
\begin{equation}
    \mathrm{AdaSoftMax}(x) = \mathrm{Softmax}(x/T).
    \vspace{-5pt}
\end{equation}
The adaptive-temperature $\mathrm{Softmax(\cdot)}$ facilitates the training because it can automatically adjust the uncertainty of the prediction. At the beginning of the training, the prediction uncertainty is high and $T$ is adjusted to a large value. At the end of the training, the prediction uncertainty is low and $T$ is adjusted to a small value.

Assume $\mathbf{Y}_{k}$ is the predicted semantic edge probability of label $k$, $\overline{\mathbf{Y}}_{k}$ is the binary semantic edge groundtruth label of class $k$, $\mathbf{p}$ represents pixel location, and $\gamma$ is the non-edge pixel percentage
\begin{equation}\label{eq:loss}
    \begin{split}
        \mathbf{\hat{Y}}  &= \mathrm{AdaSoftMax}(\mathbf{Y}),\\
        \mathcal{L}_s  &=  \sum_{k} \sum_{\mathbf{p}}\left\{-\gamma \overline{\mathbf{Y}}_{k}(\mathbf{p}) \log \mathbf{\hat{Y}}_{k}(\mathbf{p} )\right.\\
&\left.-(1-\gamma)\left(1-\overline{\mathbf{Y}}_{k}(\mathbf{p})\right) \log \left(1-\mathbf{\hat{Y}}_{k}(\mathbf{p}  )\right)\right\}.
    \end{split}
\end{equation}
For instance center regression and the instance offset regression, we use the $L_2$ norm as the loss function for the instance center regression task as $L_{c}$ where $c_i$ is the predicted center and $\overline{c}_i$ is the groundtruth center,
and we use the $L_1$ norm as the loss function for the offset regression task as $L_{o}$ where $o_i$ is the predicted offset and $\overline{o}_i$ is the groundtruth offset
\begin{equation}
    L_{c} = \sum \Vert c_i - \overline{c}_i\Vert_2, \quad L_{o} = \sum \Vert o_i-\overline{o}_i\Vert_1.
\end{equation}
The final loss function is
\begin{equation}
    L = \alpha_s L_{s} + \alpha_c L_{c} + \alpha_o L_{o}.
\end{equation}

\section{Experiments}~\label{sec:exepriements}
In this section, we present the qualitative and quantitative results of the proposed method.
We evaluated our method on the Cityscapes~\cite{cordts2016cityscapes} dataset. The splits of training, validation and testing samples are $2975$, $500$, and $1525$. The dataset contains $8$ "thing" classes and $11$ "stuff" classes. We convert the original panoptic segmentation mask annotation to panoptic edge annotation by applying erosion and exclusive OR (XOR) operator $\oplus$ from the panoptic segmentation label. Assuming the original panoptic segmentation label is $\mathbf{P}$, we can obtain panoptic edge label $\mathbf{PE}$ as follows where $r$ is the width of the edges. In this work, we try to tackle thin edge setting $r=2$.
\vspace{-5pt}
\begin{equation}
   \mathbf{PE} = \mathrm{erosion}(\mathbf{P}, r) \oplus \mathbf{P}.
   \vspace{-5pt}
\end{equation}
We implement our pipeline based on Detectron2~\cite{wu2019detectron2} and PyTorch~\cite{paszke2019pytorch}.We use Adam\cite{kingma2014adam} optimizer with the base learning rate as $0.001$. We train our model with $90000$ iterations based on ResNet$-$$18/50$ model~\cite{he2016deep}.

\subsection{Evaluation Protocol}
We evaluate our method using protocols similar to panoptic segmentation.
\cite{kirillov2019panoptic} proposes a Panoptic Quality (PQ) to evaluates panoptic segmentation, and \cite{cheng2021boundary} proposes Boundary PQ that uses $\min{\mathrm{BoundaryIoU}, \mathrm{MaskIoU}}$ instead of pure $\mathrm{MaskIoU}$ used in previous work to address the boundary prediction accuracy of panoptic segmentation. In this work, we employ the Boundary PQ metric, but only use Boundary IoU in the computation instead of $\min{\mathrm{BoundaryIoU}, \mathrm{MaskIoU}}$ since we do not have any mask prediction task. 

We define the panoptic edge PQ metric as follows. The $\operatorname{IoU}_{\text{edge}}$ is the intersection over union of edge prediction and edge groundtruth. The PQ metric can be computed separately for "stuff" edges and "thing" edges, the final PQ is the average of the PQ metric of all edge segments
\begin{equation}
    \mathrm{IoU} = \frac{\mathrm {Edge}_{\text{pred}} \cap \mathrm{Edge}_{\text{gt}}}{\mathrm {Edge}_{\text{pred}} \cup \mathrm{Edge}_{\text{gt}}},
\end{equation}
\begin{equation}
    \mathrm{PQ}=\underbrace{\frac{\sum_{(p, g) \in T P} \operatorname{IoU}_{\text{edge}}(p, g)}{|T P|}}_{\text {semantic edge quality (SQ) }} \times \underbrace{\frac{|T P|}{|T P|+\frac{1}{2}|F P|+\frac{1}{2}|F N|}}_{\text {recognition quality (RQ) }}.
\end{equation}

In panoptic segmentation tasks, the IoU threshold of True Positives ($\mathrm{TP}$), False Positives ($\mathrm{FP}$), False Negatives ($\mathrm{FN}$) is $50\%$. Since the alignment between edges are more difficult than masks, we set the threshold as $10\%$. Otherwise, predictions would be hard to match the groundtruth which leads to extremly low RQ metric. 
We also evaluate the computation (FLOPS), model parameter size to validate the inference efficiency and memory footprint, in order to design an efficient and memory-friendly robotic perception model.

\begin{figure*}
\smallskip
\smallskip
\centering
\rotatebox{90}{\hspace{1.0em}Groundtruth}{\hspace{1.0em}}
\begin{subfigure}[b]{0.28\linewidth}
\includegraphics[clip,width=\linewidth]{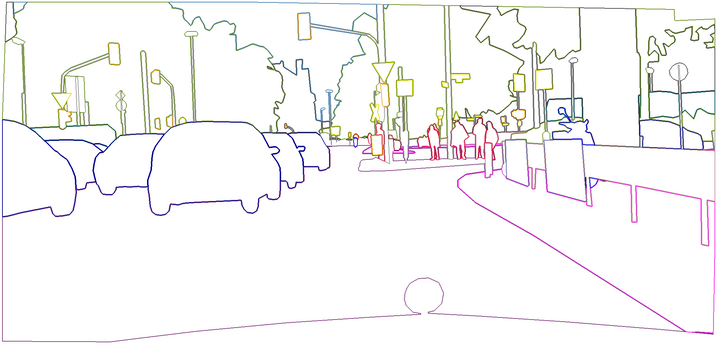}
\end{subfigure}
\begin{subfigure}[b]{0.28\linewidth}
\includegraphics[clip,width=\linewidth]{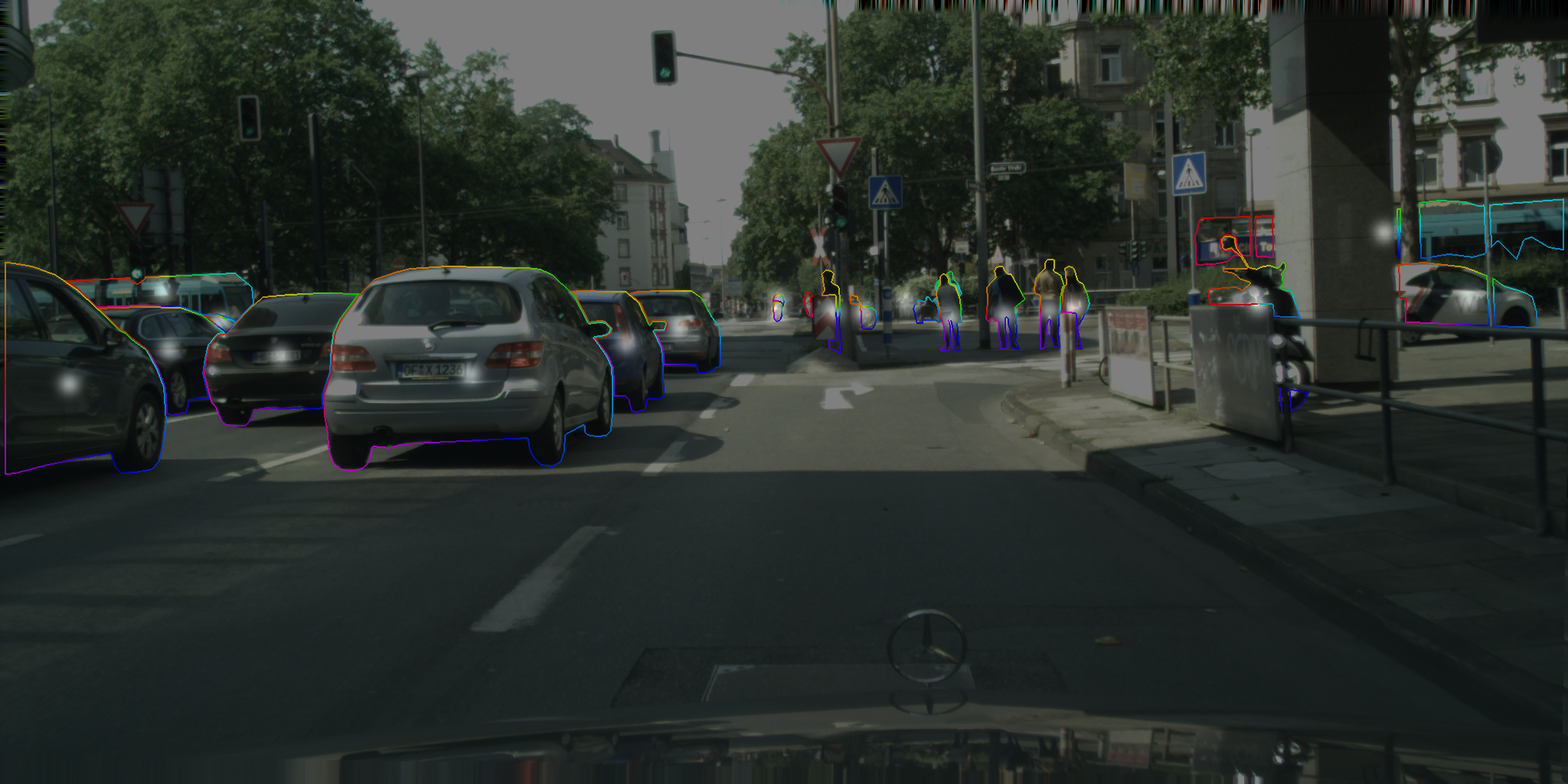}
\end{subfigure}
\begin{subfigure}[b]{0.28\linewidth}
\includegraphics[clip,width=\linewidth]{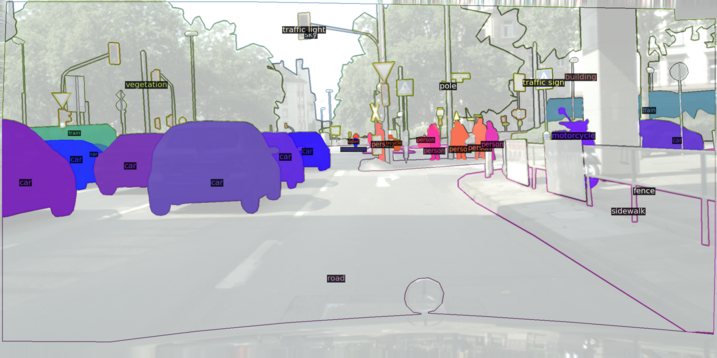}
\end{subfigure}


\rotatebox{90}{\hspace{1.0em}PENet-S}{\hspace{1.0em}}
\begin{subfigure}[b]{0.28\linewidth}
\includegraphics[clip,width=\linewidth]{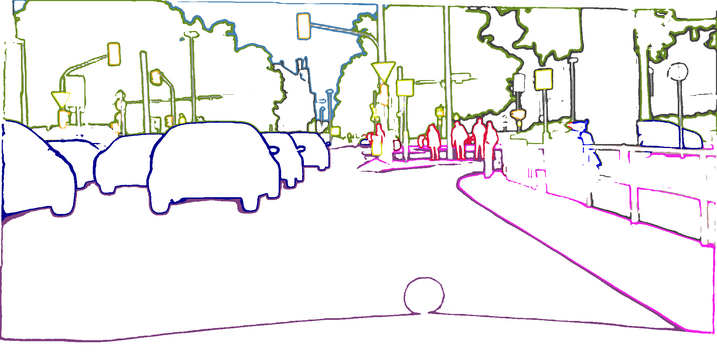}
\end{subfigure}
\begin{subfigure}[b]{0.28\linewidth}
\includegraphics[clip,width=\linewidth]{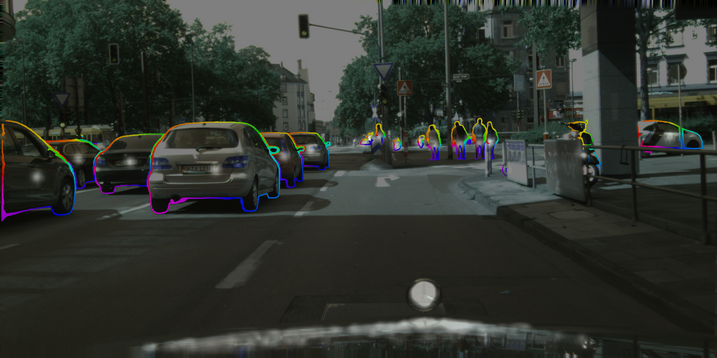}
\end{subfigure}
\begin{subfigure}[b]{0.28\linewidth}
\includegraphics[clip,width=\linewidth]{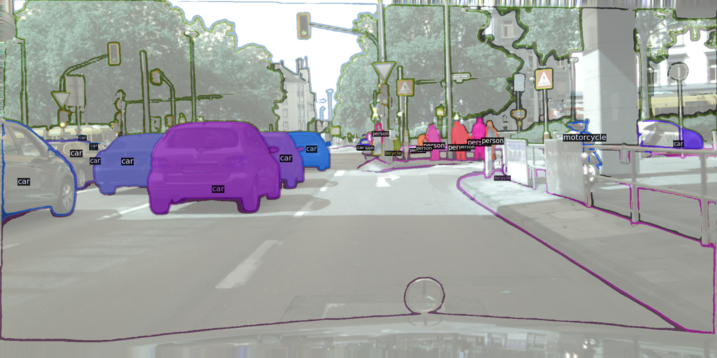}
\end{subfigure}


\rotatebox{90}{\hspace{1.0em}PENet-E}{\hspace{1.0em}}
\begin{subfigure}[b]{0.28\linewidth}
\includegraphics[clip,width=\linewidth]{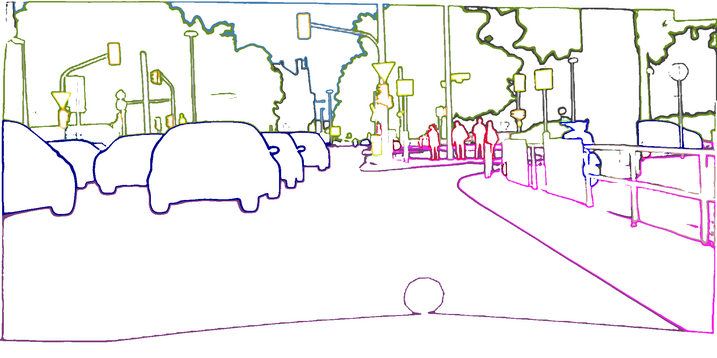}
\end{subfigure}
\begin{subfigure}[b]{0.28\linewidth}
\includegraphics[clip,width=\linewidth]{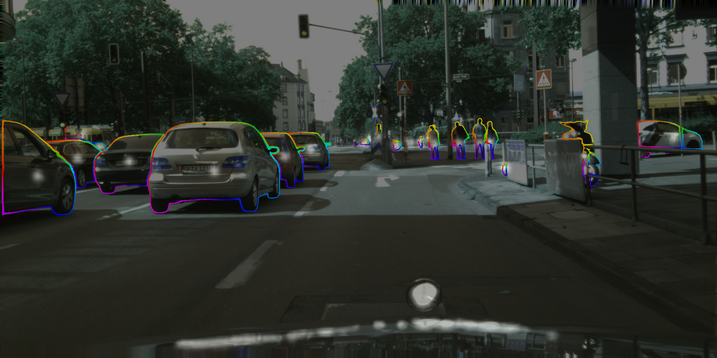}
\end{subfigure}
\begin{subfigure}[b]{0.28\linewidth}
\includegraphics[clip,width=\linewidth]{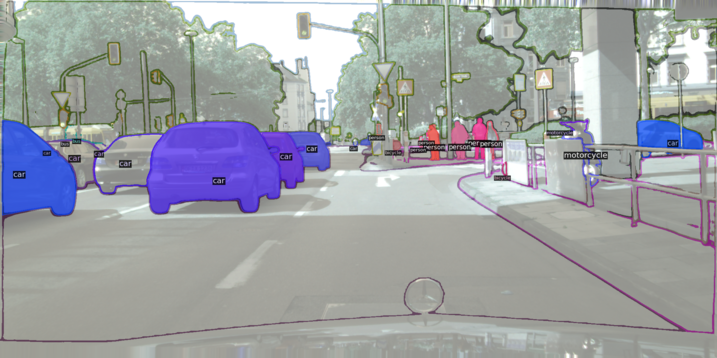}
\end{subfigure}


\rotatebox{90}{\hspace{1.0em}Groundtruth}{\hspace{1.0em}}
\begin{subfigure}[b]{0.28\linewidth}
\includegraphics[clip,width=\linewidth]{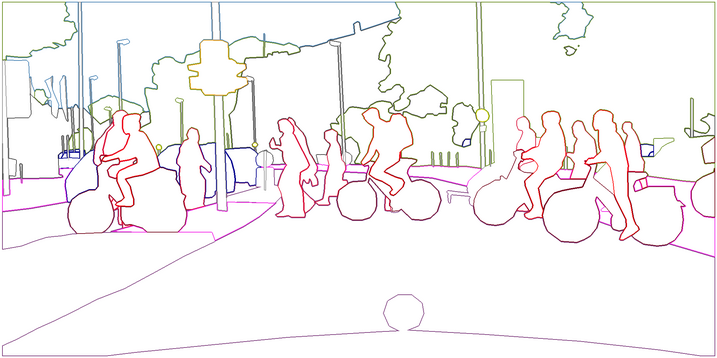}
\end{subfigure}
\begin{subfigure}[b]{0.28\linewidth}
\includegraphics[clip,width=\linewidth]{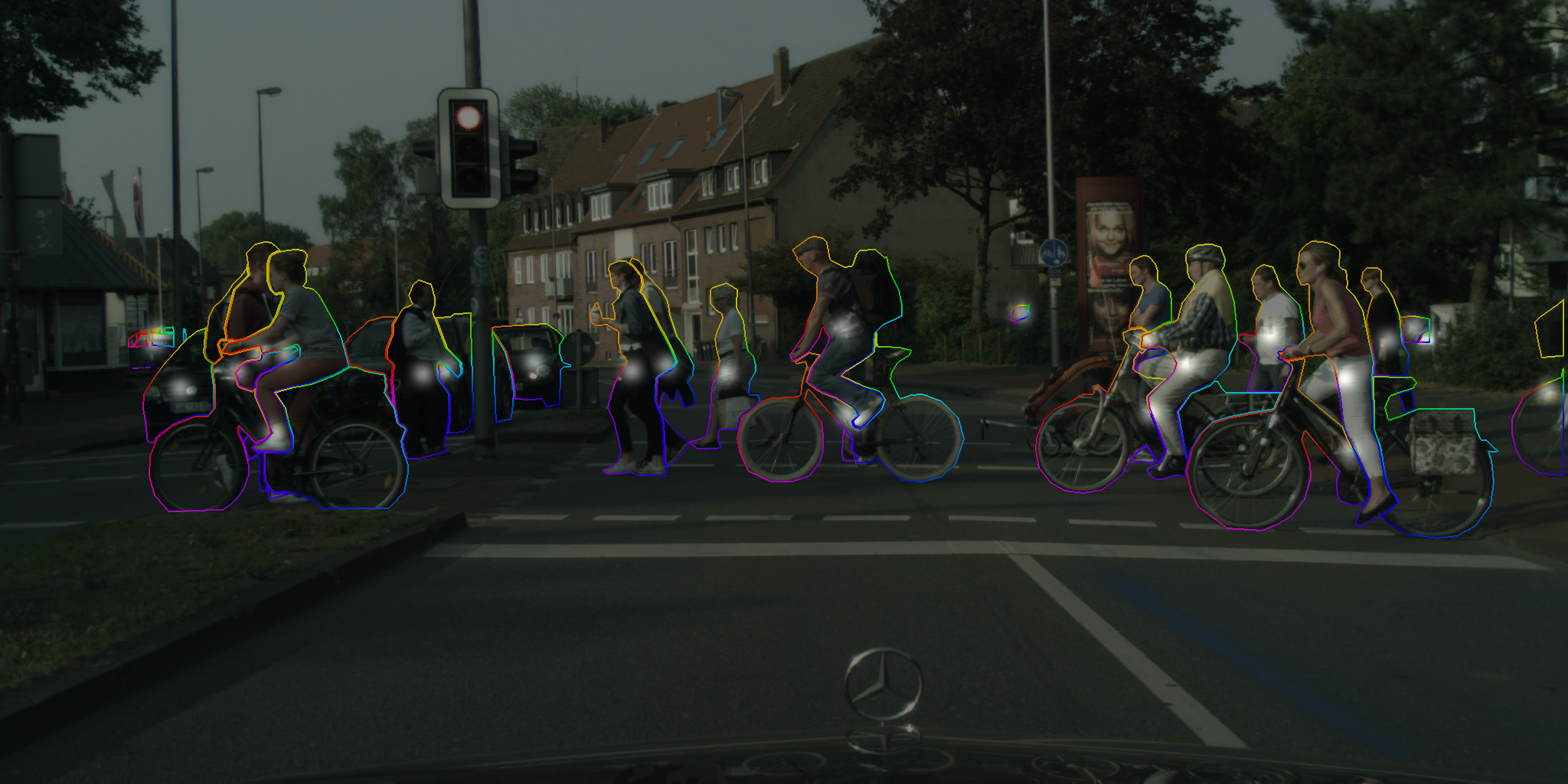}
\end{subfigure}
\begin{subfigure}[b]{0.28\linewidth}
\includegraphics[clip,width=\linewidth]{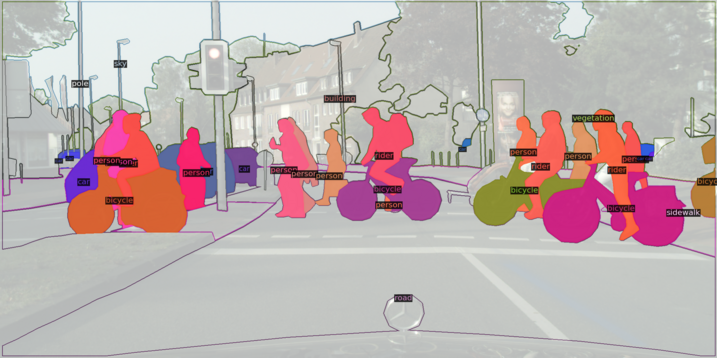}
\end{subfigure}


\rotatebox{90}{\hspace{1.0em}PENet-S}{\hspace{1.0em}}
\begin{subfigure}[b]{0.28\linewidth}
\includegraphics[clip,width=\linewidth]{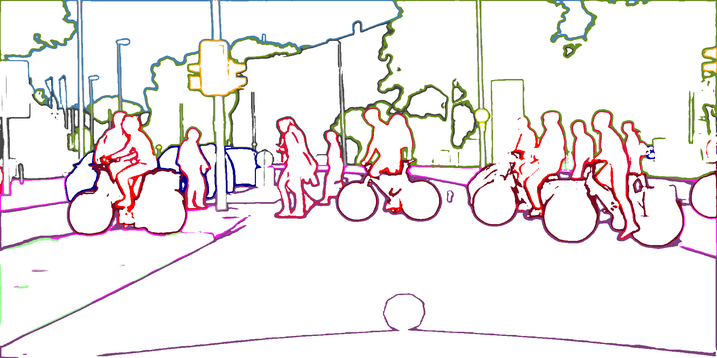}
\end{subfigure}
\begin{subfigure}[b]{0.28\linewidth}
\includegraphics[clip,width=\linewidth]{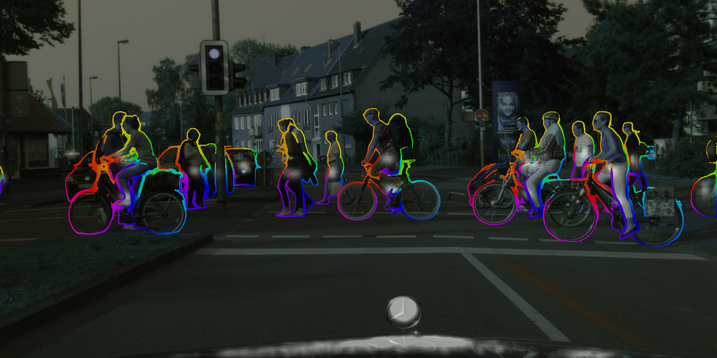}
\end{subfigure}
\begin{subfigure}[b]{0.28\linewidth}
\includegraphics[clip,width=\linewidth]{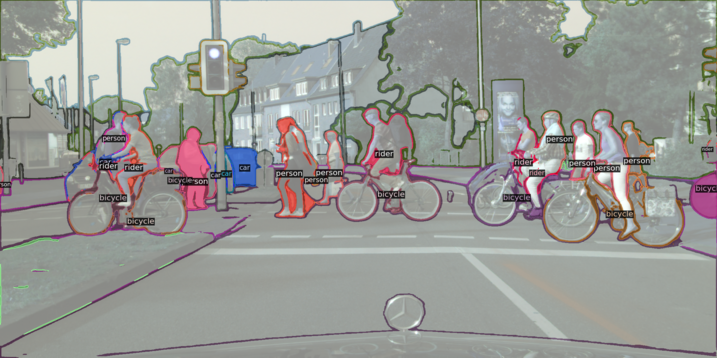}
\end{subfigure}


\rotatebox{90}{\hspace{1.0em}PENet-E}{\hspace{1.0em}}
\begin{subfigure}[b]{0.28\linewidth}
\includegraphics[clip,width=\linewidth]{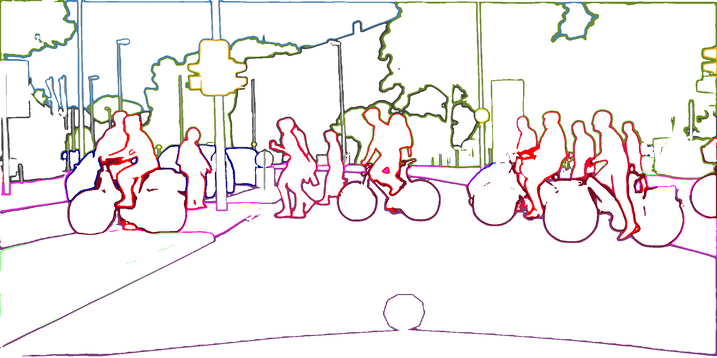}
\end{subfigure}
\begin{subfigure}[b]{0.28\linewidth}
\includegraphics[clip,width=\linewidth]{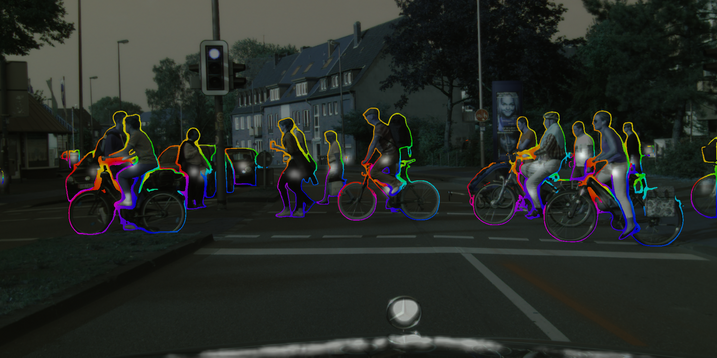}
\end{subfigure}
\begin{subfigure}[b]{0.28\linewidth}
\includegraphics[clip,width=\linewidth]{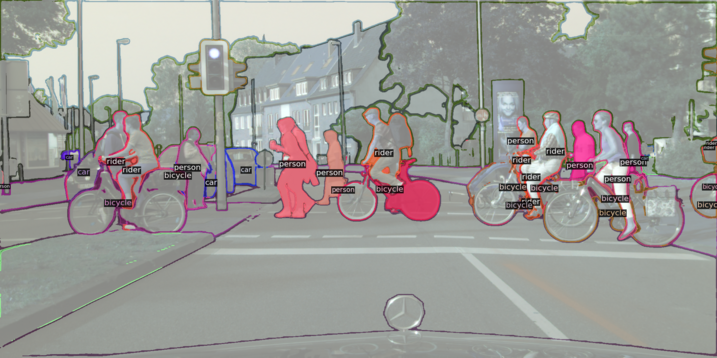}
\end{subfigure}


\rotatebox{90}{\hspace{1.0em}Groundtruth}{\hspace{1.0em}}
\begin{subfigure}[b]{0.28\linewidth}
\includegraphics[clip,width=\linewidth]{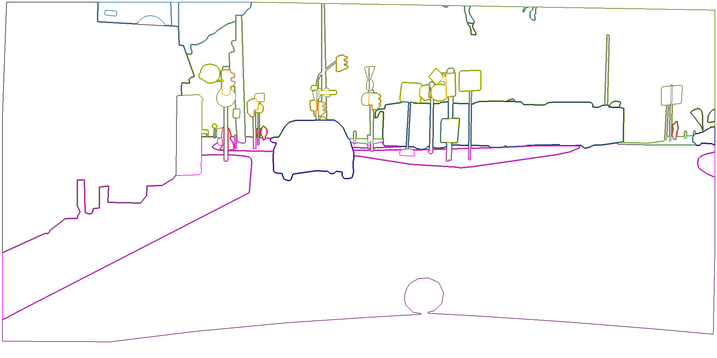}
\end{subfigure}
\begin{subfigure}[b]{0.28\linewidth}
\includegraphics[clip,width=\linewidth]{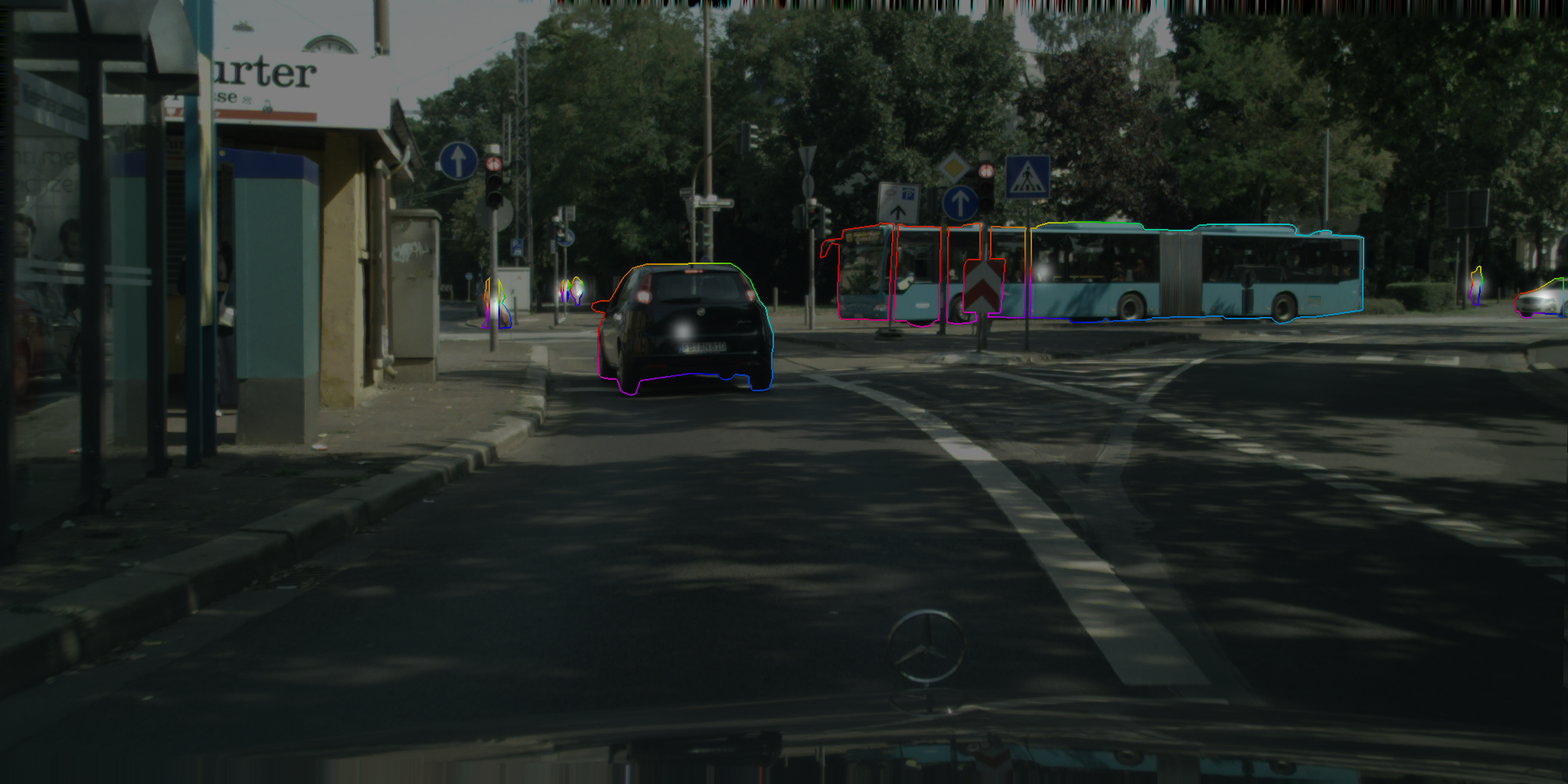}
\end{subfigure}
\begin{subfigure}[b]{0.28\linewidth}
\includegraphics[clip,width=\linewidth]{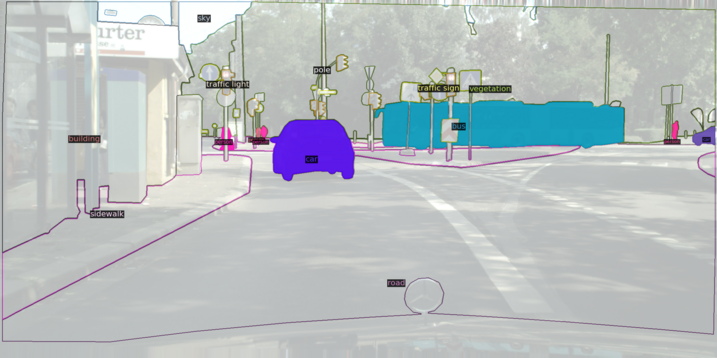}
\end{subfigure}


\rotatebox{90}{\hspace{1.0em}PENet-S}{\hspace{1.0em}}
\begin{subfigure}[b]{0.28\linewidth}
\includegraphics[clip,width=\linewidth]{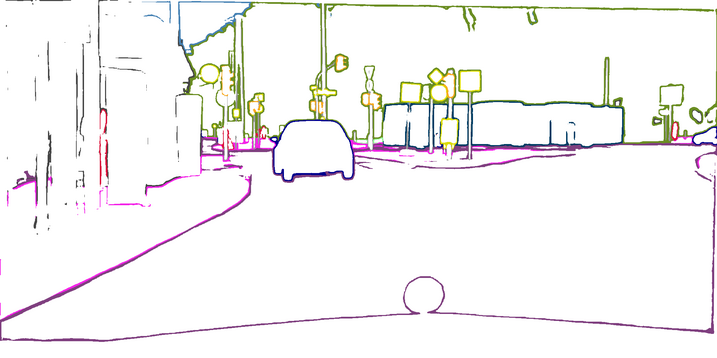}
\end{subfigure}
\begin{subfigure}[b]{0.28\linewidth}
\includegraphics[clip,width=\linewidth]{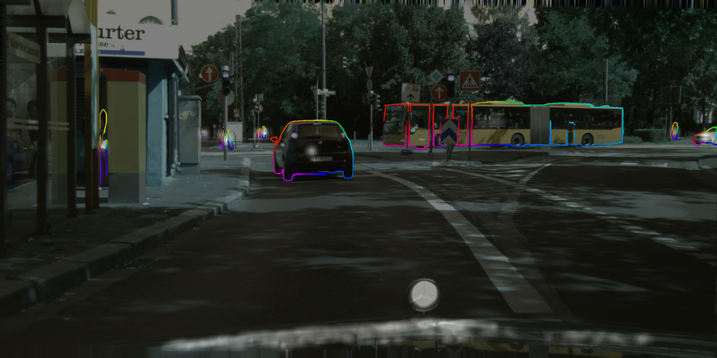}
\end{subfigure}
\begin{subfigure}[b]{0.28\linewidth}
\includegraphics[clip,width=\linewidth]{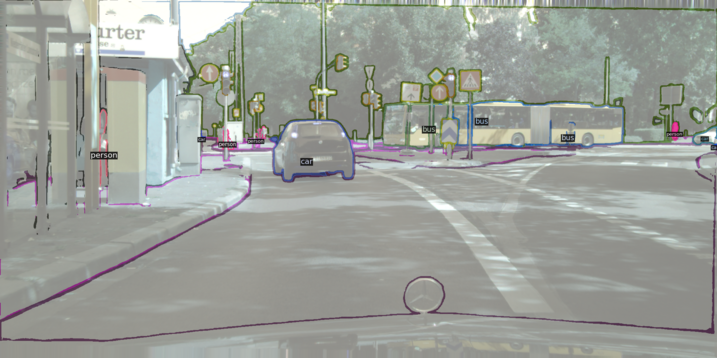}
\end{subfigure}


\rotatebox{90}{\hspace{1.0em}PENet-E}{\hspace{1.0em}}
\begin{subfigure}[b]{0.28\linewidth}
\includegraphics[clip,width=\linewidth]{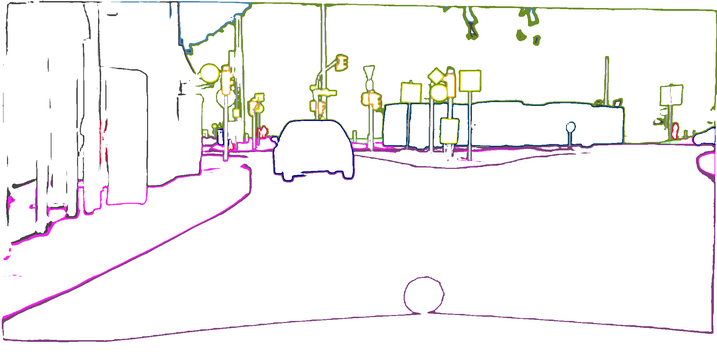}
\end{subfigure}
\begin{subfigure}[b]{0.28\linewidth}
\includegraphics[clip,width=\linewidth]{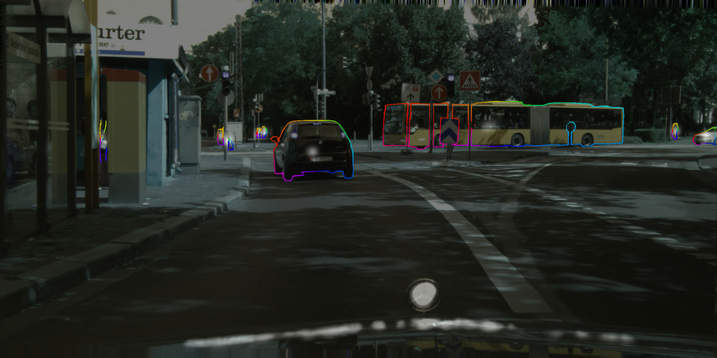}
\end{subfigure}
\begin{subfigure}[b]{0.28\linewidth}
\includegraphics[clip,width=\linewidth]{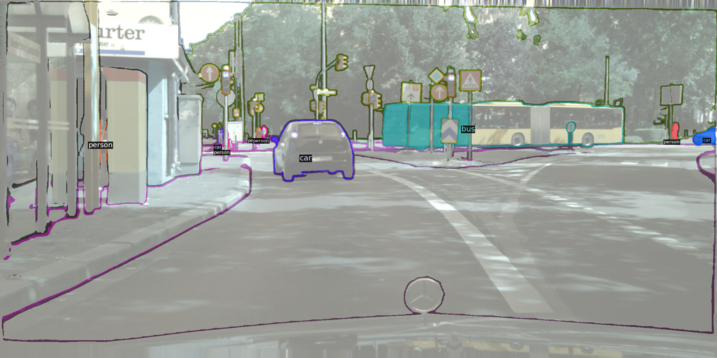}
\end{subfigure}
\caption{Qualitative Result. Columns from left to right: semantic edge, instance center heatmap with instance offset flow map, panoptic edge. We show that both PENet-S and PENet-E are robust to small objects and cluttered environments. Both qualitative results are close to the groundtruths. PENet-S performs slightly worse than PENet in the detailed areas.}
\label{fig:vis}

\end{figure*}
\subsection{Quantitative Result}
In this section, we show the quantitative result of our approach.  Since we propose a new panoptic edge definition and an evaluation metric, and the most related work \cite{Hu2019PanopticDetection} uses a different definition (multi-label edges) and metric (bounding-box based RQ metric), we cannot directly compare with this method. Therefore, we will focus on ablation study for different design choices. 

In Table~\ref{tab:main}, we show the ablation analysis of the different modules. By comparing variants PENet-A and PENet-B, we show that adopting the reweighted loss rather than the average loss of edge pixels and non-edge pixels is effective. In addition, by comparing variants PENet-B and PENet-C, we show that the output channels of the Projection Module do not affect the performance by a large margin, so we can use smaller projection channels to decrease memory usage (M: [$16$,$32$,$64$], S: [$1$,$1$,$1$]). We show by comparing PENet-B and PENet-D that Proj-B Module which utilizes criss-cross attention mechanism improves the performance against Proj-A Module and also decreases the required FLOPS computation. The variants PENet-A/B/C/D/E use ImageNet~\cite{Deng2010ImageNet:Database}-pretrained backbone model weights. The difference between PENet-D  and PENet-E  is that the semantic edge detection branch of the  PENet-E model is pretrained  for a semantic detection task before the joint training. We can see that the PENet-E model outperforms all other models and becomes the best model of our approach. Additionally, we propose a variant PENet-S with ResNet-18 backbone that consumes $34\%$ less computation and $31.85\%$ memory size while maintaining competitive performance against other variants. We also show the detailed PQ metric of "thing" and "stuff" edges separately in Table~\ref{tab:sep_pq} for reference.
\vspace{-7pt}
\subsection{Qualitative Result}
In this section, we show the qualitative result of different samples on our best model PENet-E and small mobile-friendly model PENet-S. Our method has been proven to be robust against different small instance objects, environment-cluttered conditions and overlapping cases. In Fig.~\ref{fig:vis}, we show semantic edge, instance center heatmap with offset flow map, and the final panoptic edge. 
We overlay the image on the last two columns for easier recognition of edge alignments. We fill in the complete edge contours to visualize the completeness of the panoptic edge prediction. We observe that small model PENet-S tends to predict opened instance contours compared to PENet-E. Both models works well on instance center and offset prediction tasks. The bottleneck of the final panoptic edge detection is the performance of semantic edge detection, especially for edges in the cluttered area. From some small area, we do not expect the performance to be perfect since some categories and instances of the edges cannot even be recognized by human.

\section{Conclusion}
In this work, we proposed PENet, a novel learning-based panoptic edge detection approach. We formulate a novel panoptic edge definition and evaluation metric and we design the approach as a multi-task joint network to exploit the cross-correlation information flow acorss the tasks. Our method is accurate and memory-friendly, and we evaluate our method on the real-world Cityscapes dataset and demonstrate its effectiveness. Future works will investigate how the proposed method can be effectively applied and be useful to multiple downstream robotic tasks to improve the situation awareness and safety of robotic operations.
\vspace{-5pt}
\section*{Acknowledgement}
This work was supported in part through the NYU IT HPC resources, services, and staff expertise.
\vspace{-5pt}

\bibliographystyle{IEEEtran} 

\end{document}